%% file: main.tex
\pdfoutput=1

\documentclass[11pt]{article}

\usepackage{EACL2023}

\usepackage{times}
\usepackage{latexsym}

\usepackage[T1]{fontenc}

\usepackage[utf8]{inputenc}

\usepackage{microtype}

\usepackage{multibib}
\usepackage{multirow}
\usepackage{tabularx}
\usepackage{soul}
\usepackage{titlesec}
\usepackage{inconsolata}
\usepackage{color}

\usepackage{algpseudocode}
\usepackage{booktabs}
\usepackage{xspace}
\usepackage{float}
\usepackage{graphicx}
\usepackage{subcaption}
\usepackage{arydshln}
\usepackage{cleveref} 
\usepackage{stfloats}

\pdfpageattr {/Group << /S /Transparency /I true /CS /DeviceRGBA>>}

\newcommand{\projectName}{Computer Science Insights\xspace}
\newcommand{\projectAcronym}{CS-Insights\xspace}
\newcommand{\nlpScholar}{NLP Scholar\xspace}
\newcommand{\nlpExplorer}{NLPExplorer\xspace}
\newcommand{\googleScholar}{Google Scholar\xspace}
\newcommand{\semanticScholar}{Semantic Scholar\xspace}
\newcommand{\aclAnthology}{ACL Anthology\xspace}

\newcommand{\wos}{Web of Science\xspace}
\newcommand{\scopus}{SCOPUS\xspace}
\newcommand{\dblp}{DBLP\xspace}
\newcommand{\diii}{D3\xspace}

\title{CS-Insights: A System for Analyzing Computer Science Research}

\author{Terry Ruas\textsuperscript{1}, Jan Philip Wahle\textsuperscript{1}, Lennart K\"{u}ll\textsuperscript{2}, Saif M. Mohammad\textsuperscript{3}, Bela Gipp\textsuperscript{1} \\
        \textsuperscript{1}University of G\"{o}ttingen Germany, \textsuperscript{2}University of Wuppertal, 
        \textsuperscript{3}National Research Council Canada \\
        \textsuperscript{1}\{ruas, wahle, gipp\}@uni-goettingen.de\\
        \textsuperscript{2}leon.kuell@uni-wuppertal.de\\
        \textsuperscript{3}saif.mohammad@nrc-cnrc.gc.ca\\}
        
\begin{document}
\maketitle
\begin{abstract}
This paper presents \projectAcronym, an interactive web application to analyze computer science publications from \dblp through multiple perspectives.
The dedicated interfaces allow its users to identify trends in research activity, accessibility, author's productivity, venues, statistics, topics of interest, and the impact of computer science research on other fields.
\projectAcronym is publicly available, and its modular architecture can be easily adapted to domains other than computer science.\footnote{Demo: \url{https://youtu.be/1ryjLK7lZXA}}
\end{abstract}

\section{Introduction}
While the number of digital scientific publications keeps growing fast, our ability to analyze them does not follow the same speed, preventing us from uncovering implicit patterns among its main features (e.g., authors, venues) \cite{BornmannHM21}. 
The challenge in analyzing large amounts of articles, and possibly any type of data, comes largely from its storage and processing.
Current solutions often focus on the storage of data  (\dblp\footnote{\url{https://dblp.org}}), specific sub-areas (\nlpScholar\cite{Mohammad20a}) or dataset augmentation (\diii\cite{wahle-etal-2022-d3}).
Other robust alternatives such as \wos\footnote{\url{https://www.webofscience.com}}  and \scopus\footnote{\url{https://www.scopus.com/}} offer more complete solutions, i.e., data storage, crawling, processing, and visualization, but unfortunately lie behind paywalls, which is prohibitive to those who would benefit the most from their resources (e.g., institutions in developing countries as they usually have a restrictive budget for such services).

With more than 389 million records, Google Scholar is probably the most comprehensive academic repository today \cite{Gusenbauer19a}.
Even if not all records are publicly available, for research labs with limited funding, it is computationally unfeasible to build a tool that can store and process such massive amounts of data.
Therefore, efforts in exploring scientific publications are focused on specific niches, such as \nlpScholar \cite{Mohammad20a}, a tool to analyze natural language processing publications, or PubMed\footnote{\url{https://pubmed.ncbi.nlm.nih.gov/}\label{foot-pubmed-link}}, a system for medical sciences.
Areas without dedicated solutions rely on scientometric studies on either general data repositories (e.g., arXiv), tools (e.g., VOSViewer \cite{vanEckW10}) or (semi-)manual approaches \cite{RuasP14, SahebSC21}.

As computer science publications have been growing exponentially in the last decades \cite{WahleRMG22}, and their presence in solving or facilitating other field-related problems is undeniable (e.g., plagiarism detection \cite{WahleRFM22, WahleRKG22a}, media bias \cite{SpindePKR21, SpindeKRM22}); we see computer science as a promising environment for developing a system to help understand its publications in an automated and transparent way.

We propose \projectName (\projectAcronym), an open source\footnote{\url{https://github.com/jpwahle/cs-insights}\label{foot:csi-repo}} web-based application to retrieve and analyze computer science publications from \dblp through multiple perspectives interactively.
\projectAcronym is freely available through the project homepage\footnote{\url{https://jpwahle.com/cs-insights}\label{foot:csi-page}}.
The interactive tool enables its users to explore large amounts of data intuitively in the browser through several specialized dashboards: \textit{papers}, \textit{authors}, \textit{venues}, \textit{types of paper}, \textit{fields of study}, \textit{publishers}, \textit{citations}, and \textit{LDA topics}.
Each dashboard offers a dedicated visualization panel and eight additional filters that can be combined to investigate authors, venues, publishers, and their publications. 
This paper details how we built our interactive tool, its main components, capabilities, and some exploratory experiments to show its main functionalities.

In addition to the examples presented in \Cref{sec:experiments}, \projectAcronym can also be used to explore particular topics of interest for individual researchers and find relevant subject experts, influential publications, or important venues to inform their own research.
Conference organizers and research organizations such as ACL can use \projectAcronym to identify the community's needs, inform policy decisions, and track how the implemented interventions impact broad publication trends over time. 
For example, in NLP, one can use \projectAcronym to track how much research is performed in various fields; it can be used to track citation gaps across authors and venues; and in the future, it can be used to track the influence of big technology companies, highly-funded universities, and governments, and estimate the amount of research performed in various languages.

\section{Related work} \label{sec:related_work}
Even though tools such as \googleScholar\footnote{\url{https://scholar.google.com/}}, \semanticScholar\footnote{\url{https://www.semanticscholar.org/}}, \nlpScholar \cite{Mohammad20a}, and \nlpExplorer \cite{ParmarJJJ20} provide information on scientific documents, their use is limited.
\googleScholar and \semanticScholar focus their platforms on authors and their metrics (e.g., h-index, number of papers, citations) but lack details on venues and publishers.
Additionally, neither \googleScholar nor \semanticScholar offer an interactive, customizable platform to browse their databases, preventing users from exploring explicitly available features on their website (e.g., the field of study).
\nlpScholar and \nlpExplorer offer a more personalized solution but only focus on natural language processing (NLP) publications.
They use an interactive framework to visualize and correlate different characteristics simultaneously (e.g., venues, authors, and the field of study).
\nlpScholar offers a dynamic interface as its reports are built on top of Tableau, preventing its use as an API.
While \nlpScholar aligns the information between \aclAnthology and \googleScholar (45K articles), \nlpExplorer uses only the \aclAnthology (77K articles) as a data source, thus, limiting their use for analyses of broader trends in computer science research.

\projectAcronym offers four advantages over its competitors.
First, \projectAcronym uses \dblp, the largest collection of computer science publications, with over 6M, including \aclAnthology, arXiv, and sentient metadata (e.g., paper abstracts, author affiliations).
Second, \projectAcronym can be accessed as a REST API to retrieve the information we display in our frontend, enabling \projectAcronym to be easily incorporated into other studies.
Third, different from \wos (Clarivate) and \scopus (Elsevier), \projectAcronym is a free and transparent tool to support anyone interested in investigating publications, independently of financial status or any other ethical barrier.
Fourth, our architecture provides a scalable, customizable, and responsive service.

\section{Main components}\label{sec:ui}
\begin{figure*}[!ht]
    \centering
    \includegraphics[scale=.35]{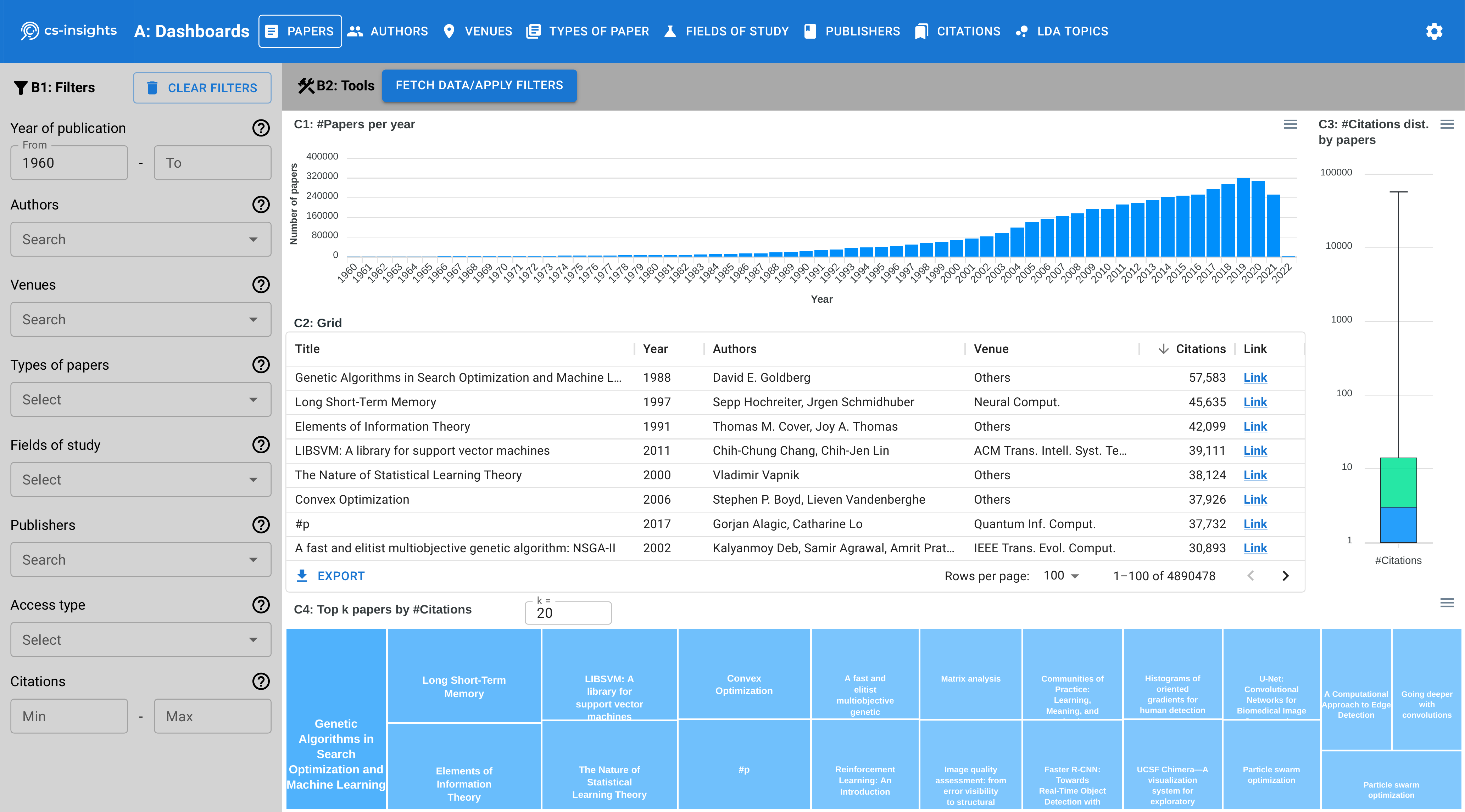}
    \caption{Frontend interface for CS-Insights.}
    \label{fig:frontend}
\end{figure*}

\projectAcronym is composed of three main boards: A. \textit{dashboards}, B. \textit{filters}, and C. \textit{visualizations}.
\textit{Dashboards} control the main views available in \projectAcronym. 
\textit{Filters} allow users to select what papers are visualized.
\textit{Visualizations} includes a series of interactive visualizations of the selected papers.
\Cref{fig:frontend} shows the landing page of our system with a default filter setting and the papers dashboard.

\subsection{Dashboards}
The \textit{dashboards} (A) are pre-set configurations focusing on specific elements of papers (e.g., number of papers, number of authors of papers, number of citations).
There are eight dashboards in total.

\textbf{Papers} shows the absolute number of papers per year; 
\textbf{Authors} shows the full name and number of authors for each paper;
\textbf{Venues} shows where each paper was published (e.g., ACL);
\textbf{Types of Papers} lists types of papers published according to their BibTex entry classification (e.g., article); 
\textbf{Fields of Study} shows the number of papers in each research area (e.g., computer science);
\textbf{Publishers} shows the responsible institution for publishing a given paper\footnote{Most publications in \dblp leave this field blank.};
\textbf{Citations} shows the accumulated incoming (i.e., how often a single paper was cited) and outgoing citations (i.e., the number of bibliography entries in a single paper);
\textbf{LDA Topics} performs a topic modeling analysis \cite{BleiNJ03} considering papers' titles and abstracts.


\subsection{Filters}
\textit{Filters} are adjustable feature-value pairs that can be configured to select a set of papers to be visualized.
There are eight filters (B1) that can be applied for each available dashboard (A); six for textual values and two for numeric ones.
The ``Fetch Data/Apply Filters'' button loads a new data batch with the selected filters (B2).

All textual filters work with auto-completion and regular expressions, thus, while typing, the user is already presented with possible matching string suggestions.
For the filters \textit{Types of papers}, \textit{Field of study}, and \textit{Access type}, pre-set values are presented in a drop-down menu.
For example, when clicking on \textit{Types of papers}, the suggested article, proceedings, book, in collection, Ph.D., and master thesis appear.
Both numerical filters (year of publication and citations) work by restricting minimum and maximum values.
Different filters work together through a logical \texttt{AND} and values on the same filter with a logical \texttt{OR}.
All filters can be cleared using the ``Clear filters'' button at the top left corner (B).
To obtain more information about the filters and their match conditions, one can hover over the question mark icon next to their heading.

\subsection{Visualizations}
There are four common visualization elements for each dashboard, as \Cref{fig:frontend} (C) shows: \textit{\#[Dashboard] per year} (C1), \textit{Paper Details Grid} (C2), \textit{\#Citations distribution} (C3), and \textit{Top $k$ by \#Citations} (C4).
The only exceptions are the \textit{Citations} and \textit{LDA Topics} dashboards, which have specific visualization elements.
The former displays \textit{Incoming} and \textit{Outgoing} citations for the selected papers as a bar chart over time, as well as a box plot for both, respectively.
The latter shows the semantic clusters and their list of frequent terms about the selected papers.
The visualization elements can be exported in several formats (e.g., csv, png).

\noindent
\textbf{\#[\textit{Dashboard}] per year} (C1) shows the amount of a given dashboard main element (e.g., papers) per year.
For example, in the \textit{Venues} dashboard, one can see a bar chart displaying the number of unique venues where the selected papers were published by year.
Hovering over a bar reveals the exact number of entries for that year.

\noindent
\textbf{Paper Details Grid} (C2) displays the available details for each dashboard choice in a table format.
For example, in the \textit{Papers} dashboard, the first column contains the paper's title followed by its year of publication, list of authors, venue, number of citations, and paper link (when available).
For the \textit{Authors} dashboard, the grid includes the name of the authors, the first and last year of publication, the number of papers, and citations.

\noindent
\textbf{\#Citations distribution} (C3) shows the distribution of citations for the selected papers.
For all dashboards, except \textit{Papers}, one can also select the number of papers as an alternative metric (B2).
Hovering over the boxplot reveals the exact values for the minimum, 25\% quartile, median, 75\% quartile, and maximum.

\noindent
\textbf{Top $k$ by \#Citations} (C4) displays the top $k$ elements based on the number of papers (regardless of publication year) in a treemap format.
As in C3, the \textit{Papers} dashboard uses the number of citations as a metric to generate its output.
All the other dashboards also offer the option of selecting the number of papers. 
The value of $k$ can be adjusted using a text field.
When the text in C4's boxes is too large, we collapse them for readability purposes.

\section{Architecture}
The \projectAcronym architecture consists of four main components\textsuperscript{\ref{foot:csi-repo}}: \textit{frontend}, \textit{backend}, \textit{prediction endpoint}, and \textit{crawler}.
Our solution is available as a free web application without the need for any installation as it runs in many web browsers\textsuperscript{\ref{foot:csi-page}}.
To guarantee a flexible and modular setup, every component in \projectAcronym runs on its own docker container. A more comprehensive list of technologies used in \projectAcronym is detailed in \Cref{ap:technologies}.

\begin{figure*}[!htbp]
    \centering
    \includegraphics[scale=.4]{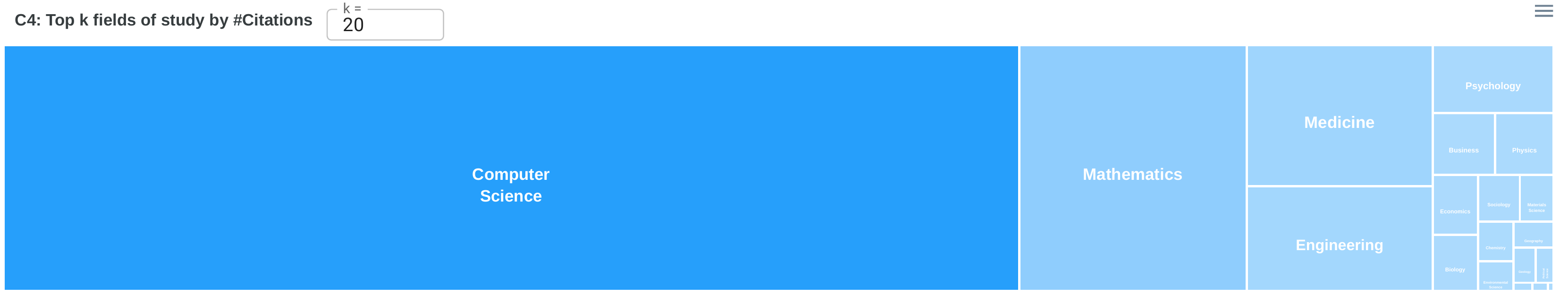}
    \caption{Top $k$ research fields by \#Citations. The size of each tile is proportional to the \#citations for that field.}
    \label{fig:fields_tree}
\end{figure*}

\textbf{Frontend.} It is responsible for presenting the main components of our tool, i.e., \textit{dashboards}, \textit{filters}, and \textit{visualizations} (\Cref{fig:frontend}).
In the frontend's interface, one can filter, retrieve, and visualize computer science publications.
We used TypeScript and React as web frameworks as they are open-access and have large community support.
For visualization, we use ApexCharts and Material UI.

\textbf{Backend.} To access, retrieve, aggregate, and analyze data, we created a REST API \textit{backend} that controls how to access data and performs computationally expensive tasks (e.g., aggregating citations of all authors for each paper available). 
\projectAcronym uses TypeScript with Node.js as JavaScript runtime and Express.js for the HTTP requests.
For persistent data storage, we use MongoDB, with mongoose providing the object document mapping.

\textbf{Prediction endpoint.} It is responsible for the training and prediction of models in the \textit{LDA Topics} dashboard used to generate the semantic topics and the lists of the most frequent and salient terms.
The project is implemented in python 3, and we use gensim's LDA \cite{BleiNJ03} implementation for topic models. 
We implement visualizations using pyLDAvis \cite{SievertS14}.
As training and inference typically require processing tens of thousands of documents, we create a dedicated service to maintain models, distribute them on the available compute infrastructure, assign them to compute jobs, and consolidate results. 

\textbf{Crawler.}
We use \dblp in our workflow as the main data source to feed \projectAcronym with computer science publications.
\dblp is currently the largest computer science repository with more than 6M documents.
To keep \projectAcronym up-to-date with the most recent publications, the \textit{crawler} downloads the latest release from \dblp, corresponding full texts, and extracts their metadata.
This pre-processing step uses the same process as in \diii \cite{wahle-etal-2022-d3}, a dataset that extends \dblp with additional information.

\section{Showcase experiments} \label{sec:experiments}
To provide an overview of \projectAcronym's core functionalities, we developed a collection of intuitive dashboards and filters that investigate broad trends in computer science publications.

\subsection{Papers overview}
\Cref{fig:frontend} shows the distribution of papers from 1960 to 2021 on the \textit{visualization} (C) board.
No filter (B) was selected to provide a high-level overview of the entire dataset.
In the \textit{\#Papers per year} visualization (C1), the number of publications constantly grew, except for the last two years.
A small decline over 2020 (and possibly 2021) can be explained by the COVID-19 pandemic that affected researchers worldwide.
The current release includes data up to December 1st, 2021\footnote{Our next data extraction will include 2021 and 2022.}.

\subsection{Fields of study}
As the number of publications has been growing rapidly, it is natural to question in which areas these publications are increasing.
Even though the number of research fields has not changed in the last couple of years (\Cref{fig:field_bar}), the number of papers between them differs greatly, as \Cref{fig:fields_tree} shows.
Computer science is overrepresented when compared to other fields, such as mathematics.

\begin{figure*}[!htb]
    \centering
    \includegraphics[scale=.7]{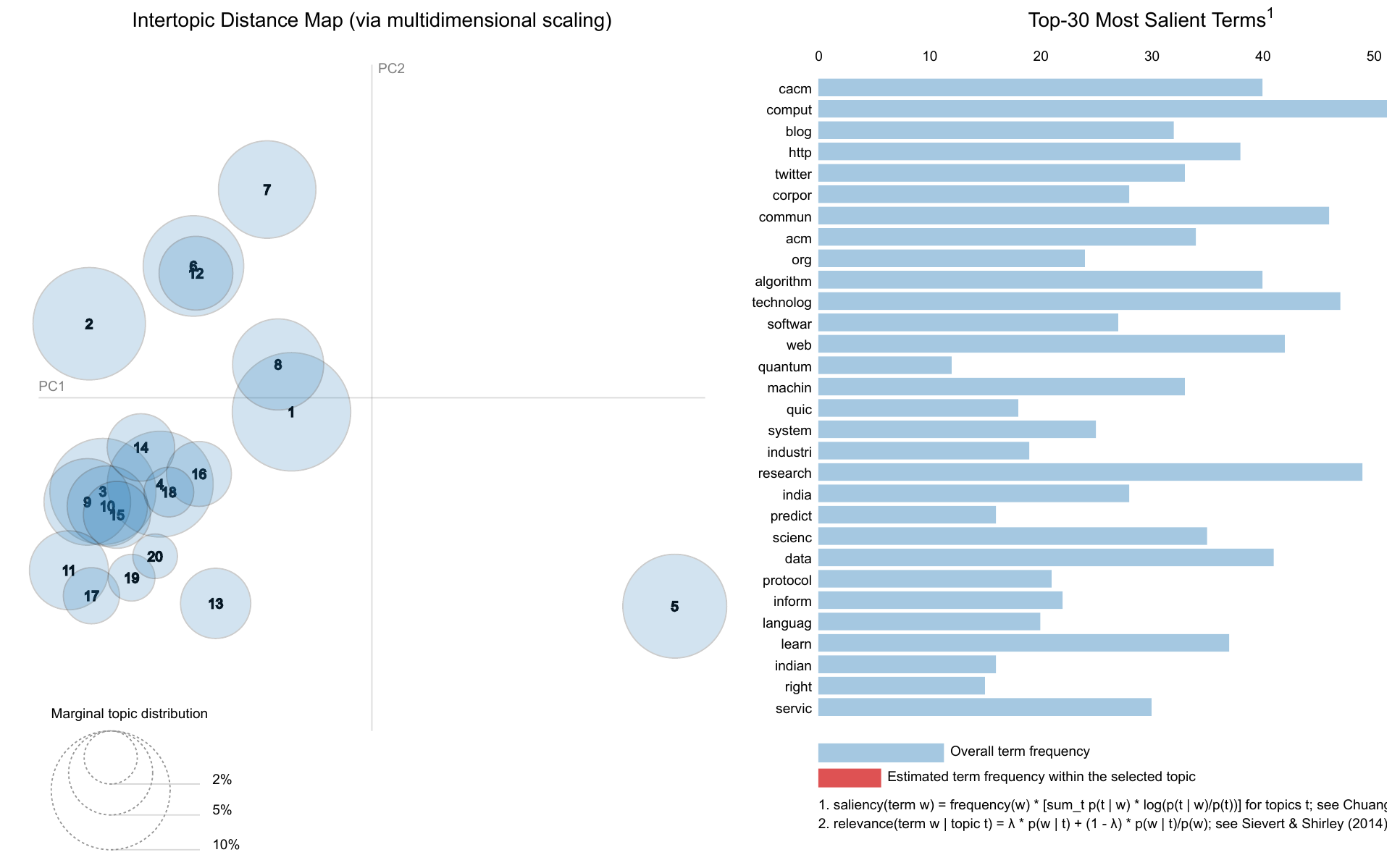}
    \caption{ (Left) The cluster of the obtained topics. Each topic is represented by a circle whose radius/size is proportional to the marginal topic distribution. (Right) LDA topics for the Communications of the ACM in 2019. The list of terms for all clusters (current state) or the terms for one specific cluster when selected.}
    \label{fig:topics_2019_full}
    \vspace*{-3mm}
\end{figure*}

In absolute numbers, computer science has approximately twice as many papers and citations as all other fields combined\footnote{Documents can be associated with more than one field.}.
In \Cref{tab:field_overview}, we show a sample for the five top fields of study with their number of papers and citations.
For a complete list of all fields of study and the boxplot of their distribution, see \Cref{tab:ap_fields} and \Cref{fig:field_box} in \Cref{sec:appendix}.

\input{tables/field_of_study}

\subsection{LDA topics}
In the \textit{LDA topics} dashboard, users can explore the most frequent and salient \cite{ChuangMH12} terms (stemmed words) of a given collection of documents through an LDA implementation for topic modeling \cite{SievertS14}.
\Cref{fig:topics_2019_full} shows the topic distribution for the ``Communications of the ACM'' venue in 2019.
The output in this dashboard is divided into the semantic cluster topics (left) and the list of the most frequent and salient terms (right).
Both parts are produced based on the text in the titles and abstracts of papers, which are parsed, stemmed, and cleaned\footnote{\url{https://radimrehurek.com/gensim/parsing/preprocessing.html}} (e.g., stopword removal, punctuation removal, punctuation removal).
In \Cref{fig:topics_2019_full}, we see an overlap of clusters, indicating their semantic proximity.
We also include the same experiment for the years 1999 (\Cref{fig:topics_1999_full}) and 2009 (\Cref{fig:topics_2009_full}) for comparison in the \Cref{sec:appendix}.
When no cluster is selected, the plots consider all titles and abstracts to compose their list of terms.
When hovering over a cluster, the 30 most relevant terms of the selected cluster are shown on the right as red bars while continuing to show the overall frequency of those 30 terms in all clusters as blue bars.
One can also identify clusters associated with a term by clicking on the desired terms directly.

\subsection{Additional experiments}\label{sec:add_exp}

In the following, we provide additional examples of analyses that can be performed with \projectAcronym.
The analytical capabilities of the tool reach far beyond the portraited examples, and with active development, more features will become available to answer more questions about CS research.
For space reasons, the tables and figures are included in the \Cref{sec:appendix}, and further analysis is included in \cite{Kull22}.


\textbf{\Cref{tab:topics_venues_authors}} states the top 30 most salient terms for the top 5 most cited authors (combined) and venues (individual) in the field. This table provides insight into the research areas and topics that have been most prominent among the most cited authors and venues in the field. For example, the terms in this table reveal that the topics related to ``image'', ``segmentation'', or ``object'' ``detection'' are the most prominent among the computer vision conference CVPR.

\textbf{\Cref{tab:author_prod}} shows the number of papers and number of citations of the top-5 most cited authors ranked by the number of papers and number of citations. This table provides information about the most cited authors in the field and how their work has been received by the research community. The data of this CS-Insights export reveals which authors have published the most papers and received the most citations and how their work has been received by the research community. For example, among all CS researchers, Ross B. Girshick (a computer vision researcher at Meta AI) has received the most citations (both on average and in total), and one of his close collaborators, Kaiming He (also at Meta AI), comes right after him.

\textbf{\Cref{fig:venues_paper}} visualizes venues considering their total number of papers. The grid in this figure provides insight into the conferences and journals with the most relevant papers in the field, e.g., which venues are most active and have the highest impact and how long they have been taking place.

\textbf{\Cref{fig:venues_citation}} illustrated venues considering their total number of citations. This figure provides insight into the conferences and journals that have received the most citations in the field. By looking at the data in this figure, it is possible to see which venues have received the most citations and how citations are generally distributed.

\textbf{\Cref{fig:topics_2019_acm_cluster}} plots the LDA topics for Communications of the ACM in 2019 with cluster 1 selected. This figure provides insight into the topics that have been discussed in Communications of the ACM in 2019. The most salient terms are, for example, ``latency'' and ``quick'', or web ``protocols'' like ``tcp'' and ``http''.

\textbf{\Cref{fig:topics_2019_acm_word}} again shows the LDA topics for Communications of the ACM in 2019 with the term ``comput'' selected. This figure additionally provides insights into the topics related to the term ``comput'' over \Cref{fig:topics_2019_acm_cluster}. Not only does the distribution of clusters change in accordance with the selected keyword, but also the order of terms changes. For example, ``http'' and ``twitter'' achieve much higher relative relevance when ``comput'' is selected.

\section{Conclusion}
We presented \projectAcronym, an interactive, open-source, and web-based visualization tool to facilitate the exploration of computer science publications.
Our tool crawls and processes papers from \dblp in a modular architecture, facilitating the maintenance and incorporation of more efficient components in the future.
In future work, we plan to incorporate improvements to our tools, such as the visualization of authors' affiliations and their correlation with existing \textit{dashboards} and collaboration graph visualizations to spot authors' and institutions' collaborations easily.
The project is currently actively developed by four contributors on GitHub, with new features shipping every month.
The currently planned features are available through the project roadmap\footnote{\url{https://github.com/users/jpwahle/projects/1/}} and currently include for example, a keyword search for papers, a split view to compare filters (e.g., authors, venues and their topics), additional impact measures (h-index, i10-index), or data expansion to 131m+ open articles on the Internet Archive Scholar\footnote{\url{https://scholar.archive.org/}} and FatCat\footnote{\url{https://fatcat.wiki/}}.

\section*{Limitations}\label{sec:limitations}
As \projectAcronym is a work in progress and it relies on external resources, there are a few limitations that should be mentioned.
Even though \dblp is the largest computer science repository, with an extensive list of features at its disposal, it does not contain all publications about computer science.

Not all characteristics available in our dataset are used for our analysis yet (e.g., the author's affiliation).
Further, some features are sparse, such as the publisher's name, which is due to missing entries in \dblp.

Due to hardware constraints, we cannot perform topic modeling for the entire dataset at runtime. 
Hence, we currently cap the number of documents to 100K that can be used for training and prediction in the \textit{LDA Topics} dashboard.
In future work, we plan to precompute common chunks of documents to allow for analyses of millions of documents.


\section*{Ethics Statement}
\projectAcronym (like other open-access analysis tools) can be misused to provide a biased and unilateral view of computer science research or related areas.
As \diii and \dblp are our data sources, all the documents used in \projectAcronym are in English; thus, a variety of other languages missing in our analysis.
Currently, our experiments and showcases hold truth for \dblp, which is an expressive subset of computer science publications, but by no means complete.
Consequently, local events and underrepresented languages are also not included in \projectAcronym.
However, researchers can use \projectAcronym to analyze trends of research about low-resource languages (including sign language) whenever it is mentioned in the title or abstract.

Another possible concern is with the non-anonymization in our visualizations.
We build \projectAcronym intending to facilitate the overview of computer science publications, their authors, venues, and topics of interest.
Therefore, we do not omit authors' names and their number of publications or citations, which can be misused.
For example, one can use available APIs, such as \textit{genderize}\footnote{\url{https://genderize.io/}}, to infer an author's gender and propagate a false correlation between productivity and gender.

The (un)intentional one-sided report of computer science publications can be used for specific political agendas.
The connection between the author's affiliations and potential cross-reference with their country can propagate a limited and unfounded view of a country or institution's scientific status.
One should be aware that there is no unique repository for all computer publications (or any other research field) worldwide. 
Hence, specific and collaborative efforts should be encouraged to obtain a more accurate perspective in our analysis. 

\projectAcronym, its components, and data are licensed to the general public under a copyright policy that allows unlimited reproduction, distribution, and hosting on any website or medium\footnote{\url{https://dblp.org/db/about/copyright}}\textsuperscript{,}\footnote{\url{https://github.com/jpwahle/cs-insights/blob/main/LICENSE}}.
Hence, anyone accessing our tool can exploit its limitations and inherited biases to propagate and amplify societal problems.

In the retrieval and pre-processing of our data, there are a few string-parsing-matching inconsistencies (e.g., umlauts in authors' names and multiple name variations for the same author).
As \projectAcronym is an ongoing project with regular releases, we hope to address the shortcomings in the future. A roadmap and issue board are available through the project's repository for more details.

\section*{Acknowledgements}
This work was partially supported by the IFI program of the German Academic Exchange Service (DAAD) under grant no. 57515252. 
We also thank Tom Neuschulten and Alexander von Tottleben for developing the prediction endpoint.

\bibliographystyle{acl_natbib.bst}
\bibliography{anthology,custom}

\appendix

\section{Appendix}
\label{sec:appendix}

\subsection{Technologies} \label{ap:technologies}
Here we list relevant technologies used in \projectAcronym:

\textbf{Environment}
\begin{itemize}
    \item Docker - \url{https://www.docker.com}
\end{itemize}

\textbf{Frontend}
\begin{itemize}
    \item TypeScript - \url{https://www.typescriptlang.org}
    \item React - \url{https://reactjs.org/}
    \item ApexCharts - \url{https://apexcharts.com/react-chart-demos/}
    \item Material UI - \url{https://mui.com/}
\end{itemize}

\textbf{Backend}
\begin{itemize}
    \item MongoDB - \url{https://www.mongodb.com}
    \item mongoose - \url{https://mongoosejs.com/}
    \item Node.js - \url{https://nodejs.org/en/}
    \item TypeScript - \url{https://www.typescriptlang.org}
    \item Javbascript - \url{https://www.w3.org/standards/webdesign/script}
    \item Express.js - \url{https://expressjs.com/}
    \item express-restify-mongoose - \url{https://florianholzapfel.github.io/express-restify-mongoose/}
\end{itemize}

\textbf{Prediction endpoint}
\begin{itemize}
    \item gensim - \url{https://radimrehurek.com/gensim/models/ldamodel.html}
    \item pyLDAvis - \url{https://pyldavis.readthedocs.io/en/latest/readme.html}
\end{itemize}

\textbf{Crawler}
\begin{itemize}
    \item aiohttp - \url{https://docs.aiohttp.org/en/stable/}
    \item GROBID - \url{https://grobid.readthedocs.io/en/latest/}
\end{itemize}

\subsection{Additional Figures \& Tables}
\begin{figure*}[htb]
    \centering
    \includegraphics[width=\textwidth]{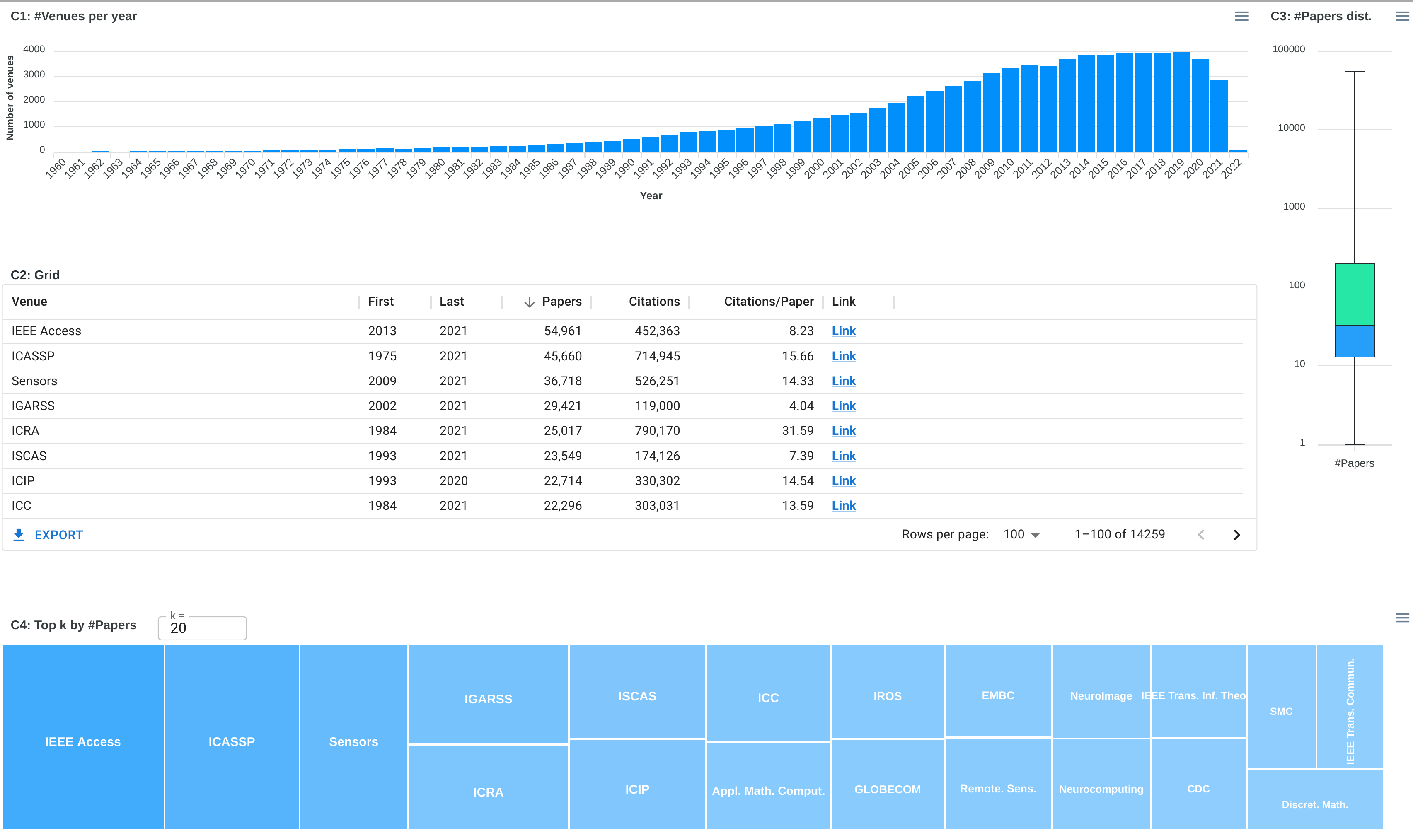}
    \caption{List of venues starting from 1960 ordered by \#papers.}
    \label{fig:venues_paper}
\end{figure*}

\begin{figure*}[htb]
    \centering
    \includegraphics[width=\textwidth]{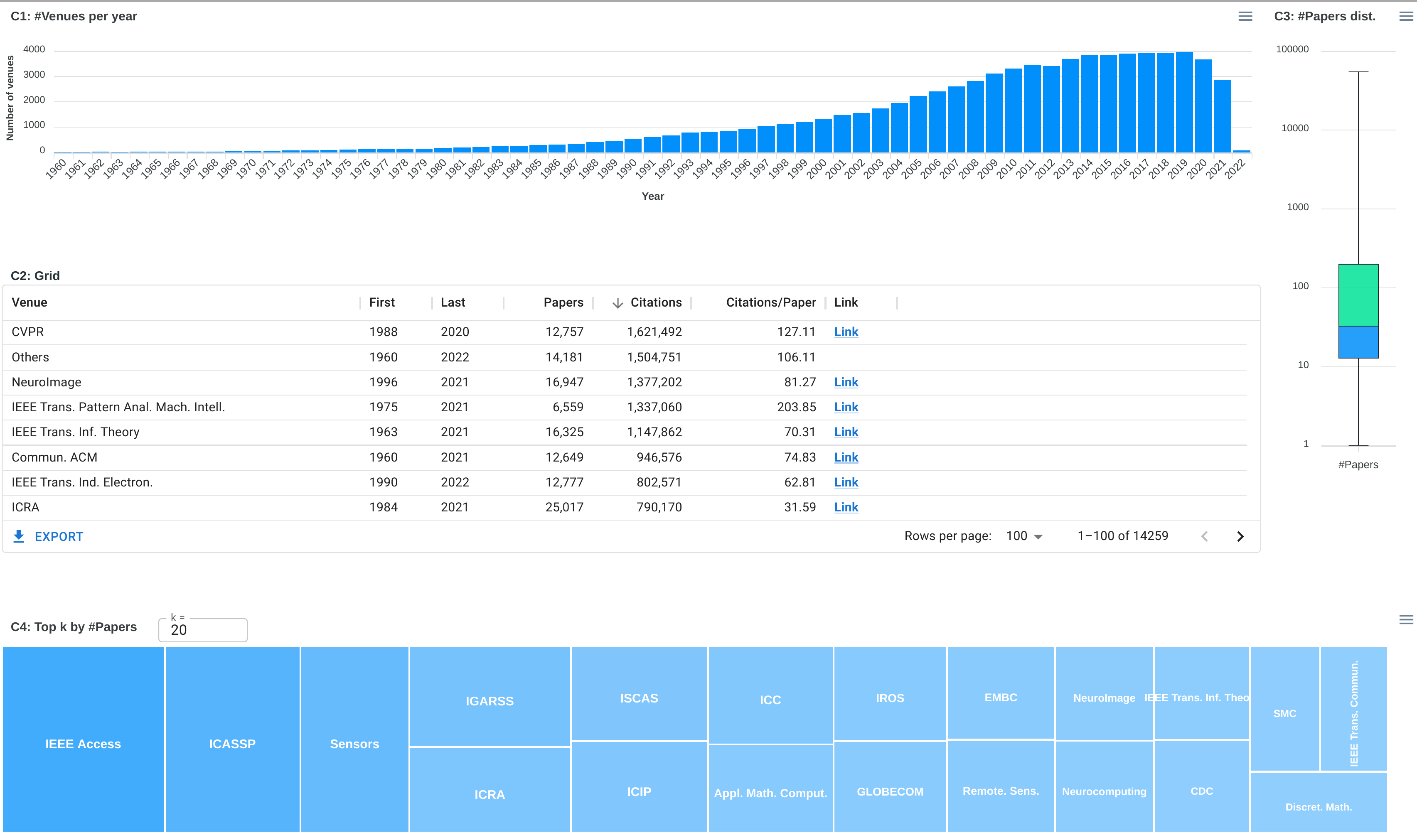}
    \caption{List of venues starting from 1960 ordered by \#citations.}
    \label{fig:venues_citation}
\end{figure*}

\begin{figure*}[htb]
    \centering
    \includegraphics[width=\textwidth]{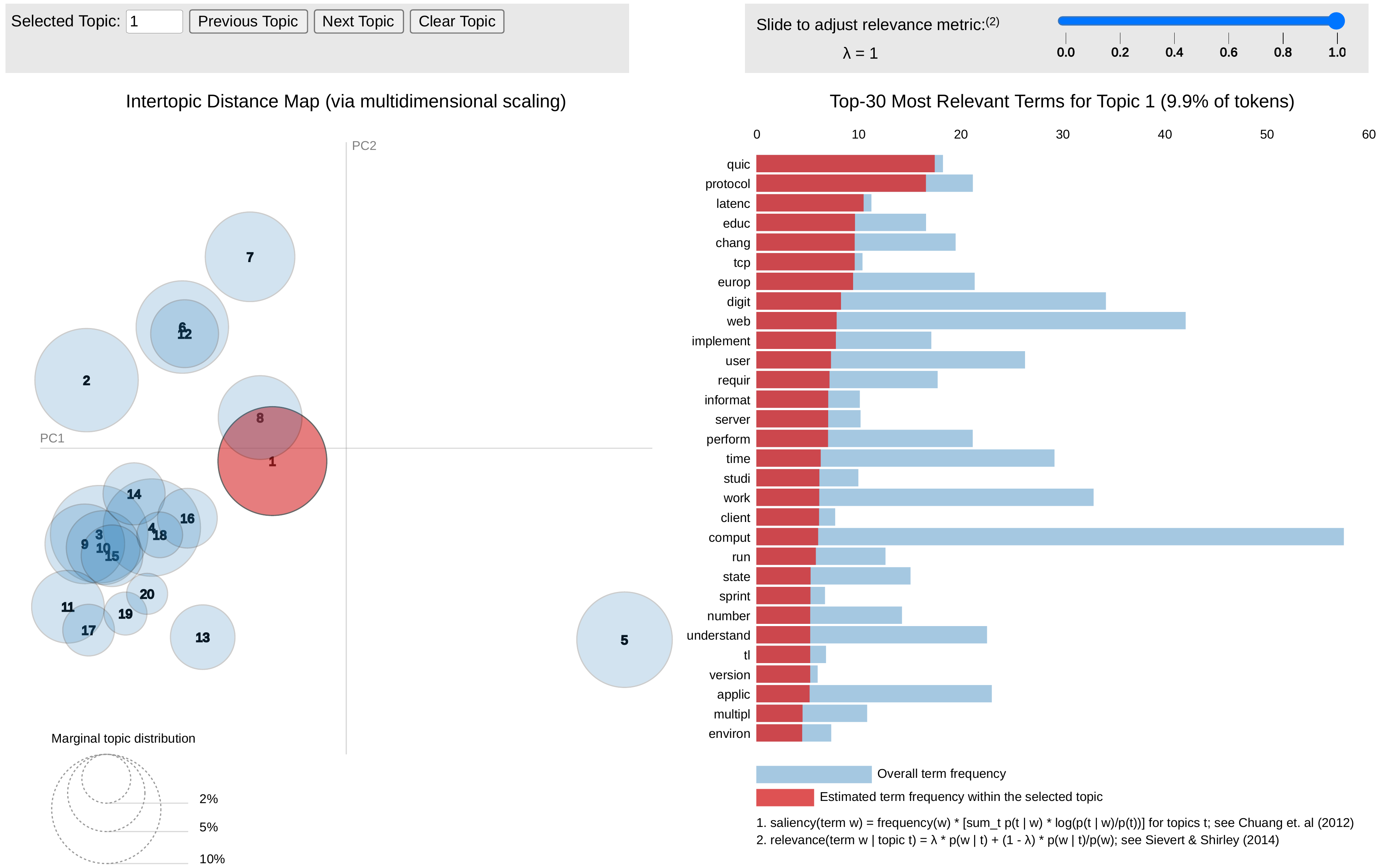}
    \caption{LDA topics for the Communications of the ACM in 2019 with cluster 1 selected.}
    
    \label{fig:topics_2019_acm_cluster}
\end{figure*}

\begin{figure*}[htb]
    \centering
    \includegraphics[width=\textwidth]{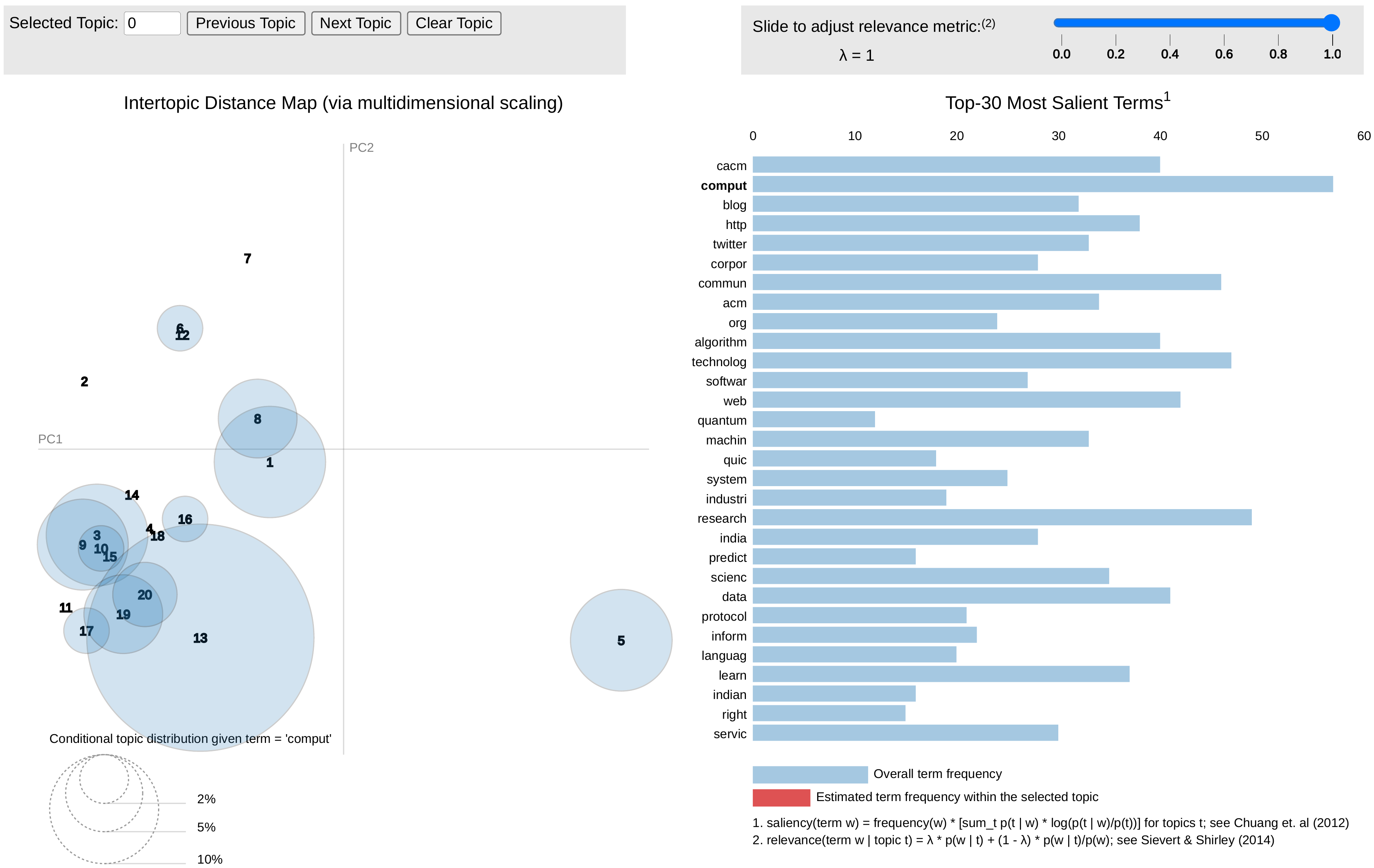}
    \caption{LDA topics for the Communications of the ACM in 2019 with the word ``comput" selected.}
    \label{fig:topics_2019_acm_word}
\end{figure*}

\Cref{tab:topics_venues_authors,tab:author_prod} and \Cref{fig:venues_paper,fig:venues_citation,fig:topics_2019_acm_cluster,fig:topics_2019_acm_word} (in the pages ahead) show example interactions with \projectAcronym that were mentioned in \Cref{sec:add_exp}.

\input{tables/topics_venues_authors} 

\input{tables/author_productivity} 

\begin{figure*}[!hb] 
    \centering
    \includegraphics[scale=0.8]{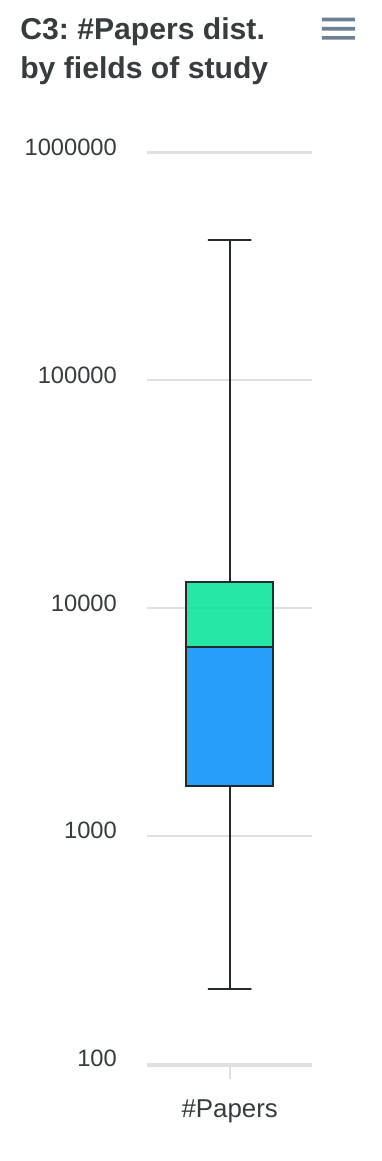}
    \caption{\#Papers distribution for the field of study.}
    \label{fig:field_box}
\end{figure*}

\begin{figure*}[!htb]
    \centering
    \includegraphics[width=\textwidth]{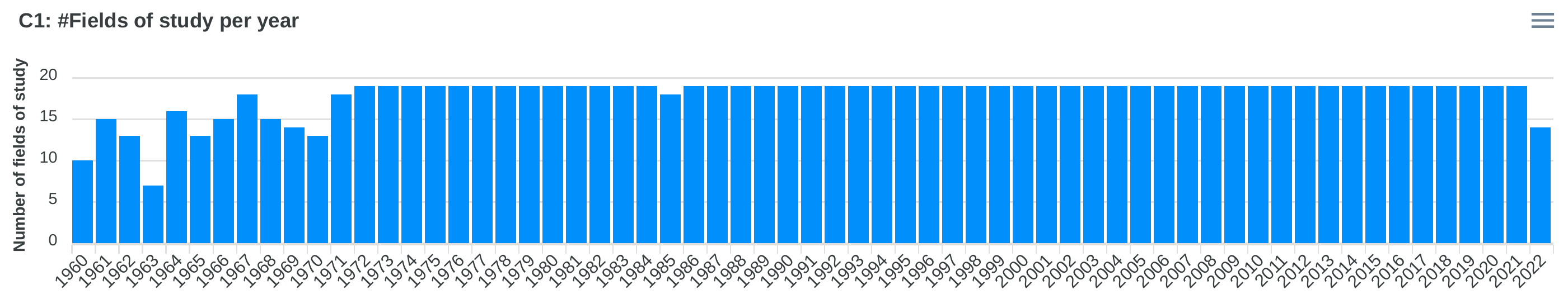}
    \caption{\#Fields of study per year.}
    \label{fig:field_bar}
\end{figure*}

\input{tables/tb_ap_fields} 

\begin{figure*}[htb] 
    \centering
    \includegraphics[scale=.75]{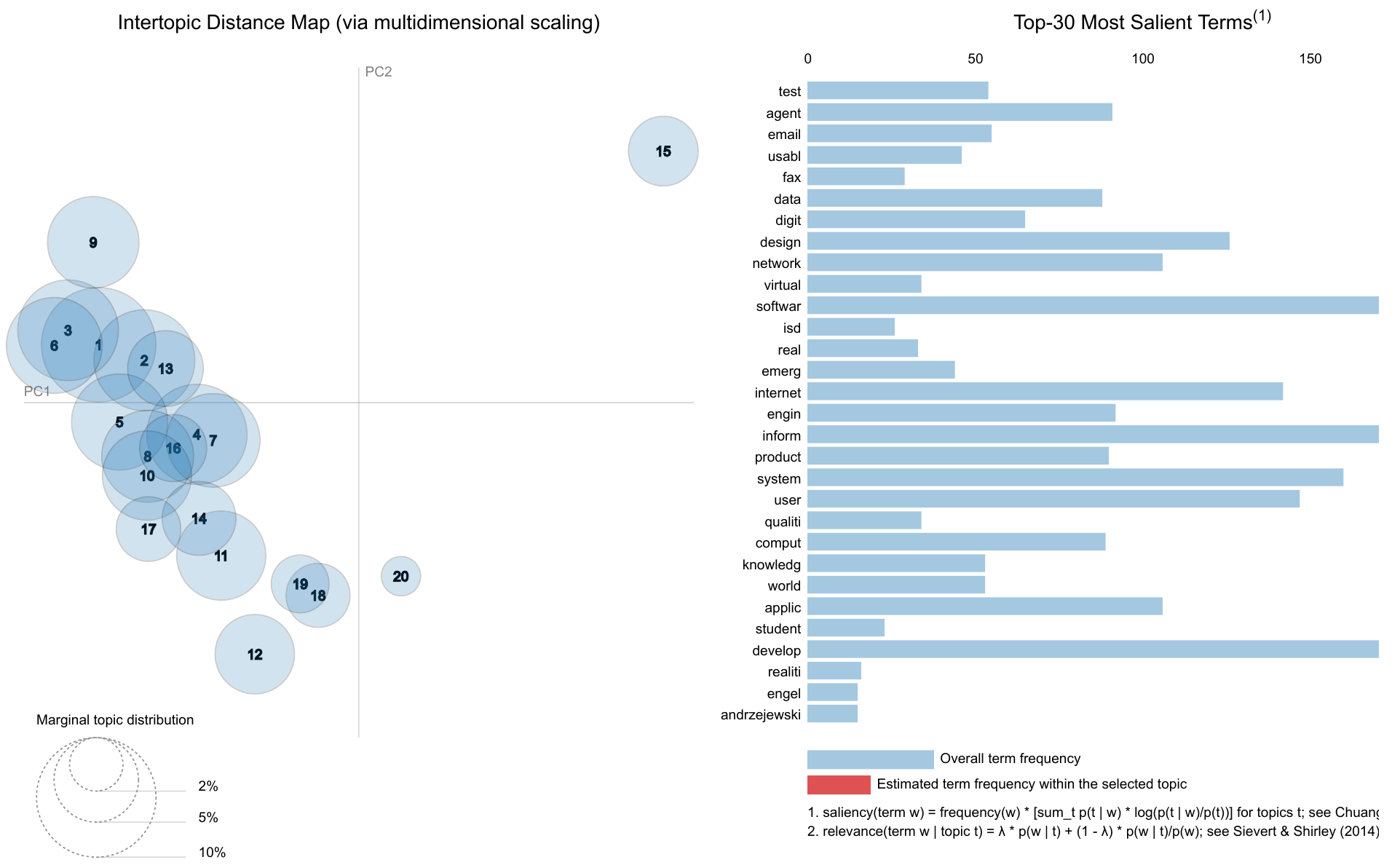}
    \caption{LDA topics for the Communications of the ACM in 1999. (Left) A cluster of the obtained topics. (Right) The list of terms for all clusters, or the terms for one specific cluster when selected.}
    \label{fig:topics_1999_full}
\end{figure*}

\begin{figure*}[!htb] 
    \centering
    \includegraphics[scale=.7]{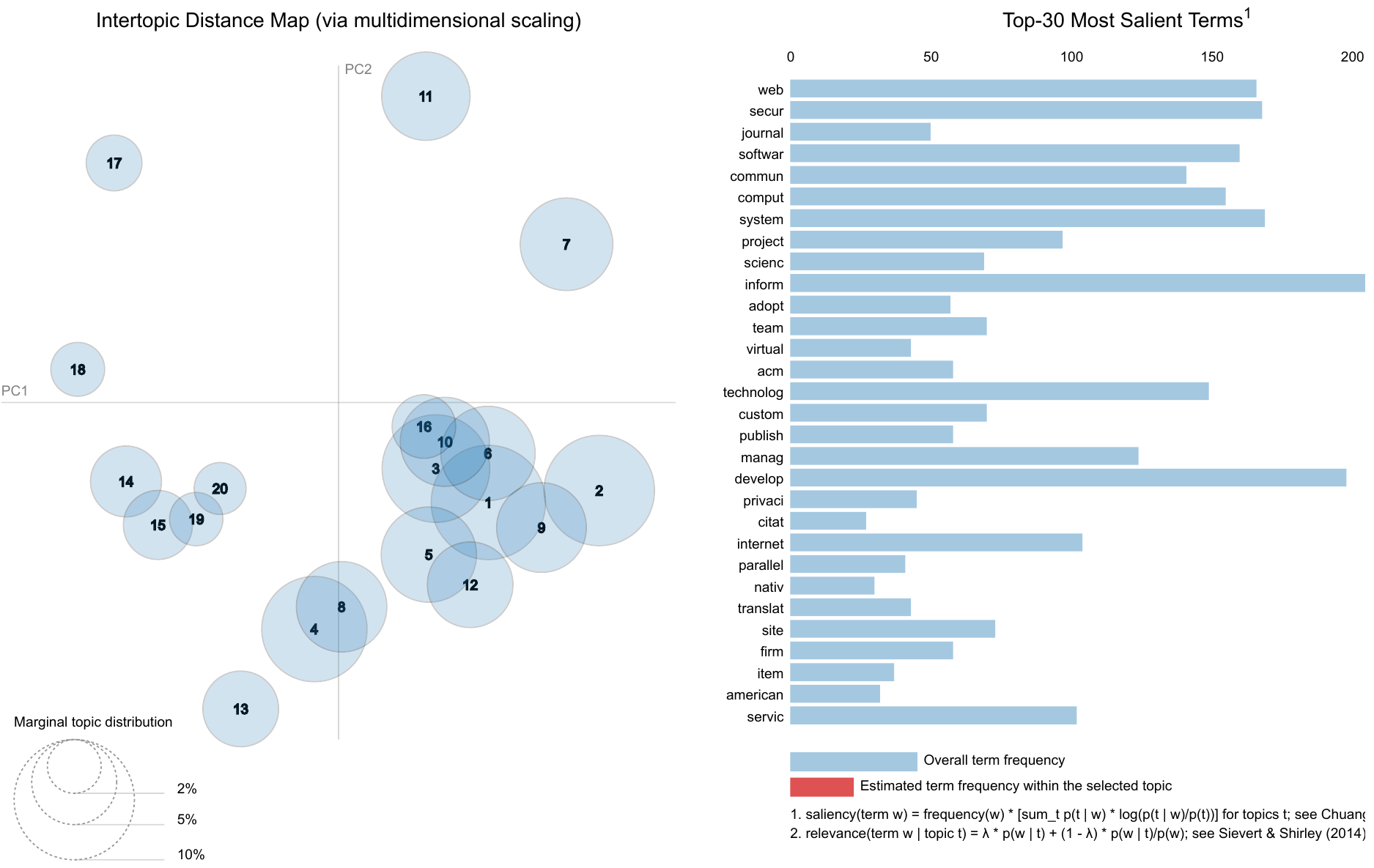}
    \caption{LDA topics for the Communications of the ACM in 2009. (Left) The cluster of the obtained topics. (Right) The list of terms for all clusters, or the terms for one specific cluster when selected.}
    \label{fig:topics_2009_full}
\end{figure*}

\end{document}

%% file: tables/field_of_study.tex
\begin{table}[t]
  \centering
  \small
  \caption{Number of papers (\#papers) and citations (\#citations) for the top five fields of study  in DBLP. }
    \begin{tabular}{lrr}
    \toprule
    \textbf{Field} & \textbf{\#Papers} & \textbf{\#Citations} \\
    \midrule
    Computer Science (CS) & 4\,189\,349 & 96\,935\,856\\
    Mathematics (MA) & 695\,351 & 21\,716\,666\\
    Medicine (MD) & 267\,794 & 10\,179\,913  \\
    Engineering (EN) & 323\,866 & 7\,550\,670\\
    Psychology (PS) & 80\,571 & 3\,199\,135\\
    \bottomrule
    \end{tabular}%
  \label{tab:field_overview}%
\end{table}%


%% file: tables/topics_venues_authors.tex
\begin{table*}[!hb]
  \centering
  \small
  \caption{Top-30 most salient topics for top-5 most cited authors (combined) and conferences (individual). \textit{Italic} terms are only present in the author's topcis. \textbf{Bold} terms are common between authors and venues.}
    \begin{tabular}{llllll}
    \toprule
    \textbf{Cited Authors} & \textbf{CVPR} & \textbf{NeuroImage} & \textbf{IT. PA. M. Int.} & \textbf{IT. Inf. Theory} & \textbf{Com. ACM} \\
    \midrule
    \textbf{network}     & \textbf{imag}  & connect & paper & code  & program \\
    \textbf{data}        & \textbf{result} & \textbf{network} &\textbf{ propos} & channel & \textbf{data} \\
    \textbf{imag}        & \textbf{featur} & activ & \textbf{result} & bound & \textbf{algorithm} \\
    \textit{pattern}     & \textbf{network} & function & \textbf{imag}  & decod & comput \\
    \textit{mine}        & \textbf{video} & respons & model & \textbf{algorithm} & softwar \\
    \textbf{problem}     & \textbf{segment} & \textbf{imag}  & \textbf{learn} & paper & \textbf{inform} \\
    \textbf{face}        & \textbf{propos} & \textbf{data}  & object & \textbf{inform} & \textbf{problem} \\
    \textbf{cluster}     & detect & \textbf{result} & match & sequenc & develop \\
    \textbf{object}      & \textbf{shape }& cortic & \textbf{algorithm} & \textbf{network} & languag \\
    \textbf{video}       & model & model & surfac & capac & time \\
    \textbf{propos}      & \textbf{object} & ag    & \textbf{problem} & error & \textbf{user} \\
    \textit{queri}       & method & task  & \textbf{face}  & sourc & research \\
    \textbf{segment}     & pose  & visual & \textbf{segment} & function & system \\
    \textbf{inform}      & local & area  & recognit & signal & number \\
    \textbf{user}        & state & method & \textbf{featur} & estim & commun \\
    \textbf{train}       & \textbf{learn} & brain & \textbf{data}  & rate  & provid \\
    \textbf{result}      & point & process & \textbf{shape} & construct & new \\
    \textbf{shape}       & \textbf{face } & matter & structur & given & acm \\
    \textbf{algorithm}   & camera & predict & motion & \textbf{problem} & \textbf{result} \\
    \textit{present}     & perform & region & estim & spl   & gener \\
    \textit{text}        & art   & cortex & \textbf{network} & sub   & \textbf{network} \\
    \textit{classif}     & \textbf{train} & state & \textbf{cluster} & time  & technolog \\
    featur      & scene & diffus & \textbf{video} & scheme & paper \\
    \textit{databas}     &\textbf{problem} & suggest & label & nois  & web \\
    \textit{graph}       & recognit & tempor & camera & multipl & \textbf{object} \\
    \textit{differ}      & track & motor & work  & model & design \\
    \textit{human}       & achiev & associ & \textbf{train} & lower & includ \\
    learn       & demonstr & eeg   & class & \textbf{result} & scienc \\
    \textit{type}        & scale & left  & function & case  & year \\
    \textit{latent}      & motion & right & requir & upper & project \\
    \midrule
     Common Topics   & 13    & 4     &16     &4   &8\\
    \bottomrule
    \end{tabular}%
  \label{tab:topics_venues_authors}%
\end{table*}%

%% file: tables/author_productivity.tex
\begin{table*}[htb]
  \centering
  \small
  \caption{Productivity and popularity for top 5 authors ranked by the number of papers and number of citations.}
    \begin{tabular}{lrrrrrr}
    \toprule
    \textbf{Rank} & \ \textbf{\#Papers} & \textbf{\#Citations} & \textbf{Year} & \textbf{Author} & \textbf{Citation avg} & \textbf{Venues} \\
    \midrule
    \multirow{5}{*}{Paper} & 1\,649  & 74\,467 & 1977  & H. Vincent Poor & 45.16 & 648 \\
          & 1\,445  & 39\,300 & 1997  & Mohamed-Slim Alouini & 27.20 & 462 \\
          & 1\,382  & 35\,887 & 1989  & Lajos Hanzo & 25.97 & 462 \\
          & 1\,287  & 73\,436 & 1980  & Philip S. Yu & 57.06 & 732 \\
          & 1\,260  & 30\,704 & 1982  & Victor C. M. Leung & 24.37 & 702 \\
    \midrule
    Average & 1\,405 & 50\,759 & 1985  &       & 35.95 & 601 \\
    \midrule
    \multirow{5}{*}{Citation} & 69    & 146\,867 & 2004  & Ross B. Girshick & 2\,128.51 & 35 \\
          & 662   & 123\,682 & 1974  & Anil K. Jain 0001 & 186.83 & 345 \\
          & 66    & 114\,330 & 2009  & Kaiming He & 1\,732.27 & 33 \\
          & 231   & 109\,821 & 1987  & Jitendra Malik & 475.42 & 131 \\
          & 454   & 105\,025 & 1985  & Andrew Zisserman & 231.33 & 250 \\
    \midrule
    Average & 296 & 119\,945 & 1992  &       & 950.87 & 159 \\
    \bottomrule
    \end{tabular}%
  \label{tab:author_prod}%
\end{table*}%

%% file: tables/tb_ap_fields.tex
\begin{table*}[!htbp]
  \centering
  \small
  \caption{Fields of study ranked by \#Papers}
    \begin{tabular}{lrrrrr}
    \toprule
    \textbf{Field} & \multicolumn{1}{l}{\textbf{\#Papers}} & \multicolumn{1}{l}{\textbf{\#Citations}} & \multicolumn{1}{l}{\textbf{First year}} & \multicolumn{1}{l}{\textbf{Last year}} & \multicolumn{1}{l}{\textbf{Avg. Citation}} \\
    \midrule
    Computer Science & 4\,189\,349 & 96\,935\,856 & 1960  & 2022  & 23.14 \\
    Mathematics & 695\,351 & 21\,716\,666 & 1960  & 2022  & 31.23 \\
    Engineering & 323\,866 & 7\,550\,670 & 1960  & 2022  & 23.31 \\
    Medicine & 267\,794 & 10\,179\,913 & 1960  & 2022  & 38.01 \\
    Psychology & 80\,571 & 3\,199\,135 & 1961  & 2022  & 39.71 \\
    Physics & 67\,156 & 1\,399\,100 & 1960  & 2022  & 20.83 \\
    Business & 57\,273 & 1\,496\,619 & 1960  & 2022  & 26.13 \\
    Materials Science & 50\,013 & 618\,573 & 1960  & 2022  & 12.37 \\
    Biology & 30\,751 & 985\,557 & 1961  & 2022  & 32.05 \\
    Economics & 30\,076 & 1\,062\,187 & 1961  & 2022  & 35.32 \\
    Sociology & 27\,714 & 762\,877 & 1962  & 2022  & 27.53 \\
    Environmental Science & 26\,354 & 410\,721 & 1964  & 2022  & 15.58 \\
    Chemistry & 17\,179 & 547\,508 & 1961  & 2021  & 31.87 \\
    Geology & 16\,926 & 304\,257 & 1962  & 2022  & 17.98 \\
    Geography & 16\,007 & 407\,229 & 1961  & 2021  & 25.44 \\
    Political Science & 15\,979 & 245\,088 & 1960  & 2022  & 15.34 \\
    Philosophy & 5\,680  & 60\,718 & 1960  & 2021  & 10.69 \\
    Art   & 4\,332  & 18\,519 & 1960  & 2021  & 4.27 \\
    History & 2\,756  & 49\,622 & 1961  & 2021  & 18.01 \\
    Others & 698\,383 & 781   & 1960  & 2022  & 0.00 \\
    \bottomrule
    \end{tabular}%
  \label{tab:ap_fields}%
\end{table*}%